\documentclass[sigconf,nonacm]{acmart}

\usepackage[inline]{enumitem}

\AtBeginDocument{%
  }

\begin{document}

\title{LITEWAY: LIghtweight HAR via Temporal Efficient highWAY}

\author{Dominique Nshimyimana}
\orcid{0009-0009-8580-1248}
\correspondingauthor
\affiliation{%
  \institution{RPTU, DFKI}
  \city{Kaiserslautern}
  \country{Germany}}
\email{dominique.nshimyimana@dfki.de}

\author{Vitor Fortes Rey}
\orcid{0000-0002-8371-2921}
\affiliation{%
  \institution{RPTU, DFKI}
  \city{Kaiserslautern}
  \country{Germany}}
\email{vitor.fortes\_rey@dfki.de}

\author{Mengxi Liu}
\orcid{0000-0003-0527-1208}
\affiliation{%
  \institution{RPTU, DFKI}
  \city{Kaiserslautern}
  \country{Germany}}
\email{mengxi.liu@dfki.de}

\author{Bo Zhou}
\orcid{0000-0002-8976-5960}
\affiliation{%
  \institution{RPTU, DFKI}
  \city{Kaiserslautern}
  \country{Germany}
}
\email{bo.zhou@dfki.de}

\author{Paul Lukowicz}
\orcid{0000-0003-0320-6656}
\affiliation{%
  \institution{RPTU, DFKI}
  \city{Kaiserslautern}
  \country{Germany}}
\email{Paul.Lukowicz@dfki.de}

\renewcommand{\shortauthors}{Nshimyimana et al.}

\begin{abstract}
Wearable human activity recognition (HAR) remains challenging due to the computational and energy constraints of deep learning models on resource-limited devices. Existing lightweight approaches often rely on recurrent architectures (e.g., GRU and LSTM), limiting parallelism and increasing inference latency. We propose \textit{LITEWAY}, a modality-agnostic, fully convolutional framework for multichannel sensor time series that replaces recurrent temporal modeling with structured convolutional decomposition. LITEWAY combines lightweight convolutional blocks, strided temporal processing, and convolution-attention pooling to efficiently capture temporal dependencies while reducing computational complexity.
We evaluate LITEWAY on 16 HAR datasets against TinyHAR, TinierHAR, and MLP-HAR. LITEWAY achieves competitive macro F1 while reducing model size by 4.06$\times$--9.52$\times$ (Light) and 3.87$\times$--9.07$\times$ (Full) compared with TinyHAR and TinierHAR. Deployment experiments further show energy reductions of 2.29$\times$--3.14$\times$ (Light) and 1.46$\times$--2.01$\times$ (Full) compared with TinierHAR and MLP-HAR, highlighting efficient fully convolutional temporal modeling for wearable HAR.
The source code is publicly available at \url{https://github.com/dominique-nshimyimana/liteway}
\end{abstract}

\begin{CCSXML}
<ccs2012>
   <concept>
       <concept_id>10003120.10003138.10003141</concept_id>
       <concept_desc>Human-centered computing~Ubiquitous and mobile devices</concept_desc>
       <concept_significance>500</concept_significance>
       </concept>
   <concept>
       <concept_id>10010520.10010553.10010562</concept_id>
       <concept_desc>Computer systems organization~Embedded systems</concept_desc>
       <concept_significance>300</concept_significance>
       </concept>
   <concept>
       <concept_id>10010147.10010257.10010293.10010294</concept_id>
       <concept_desc>Computing methodologies~Neural networks</concept_desc>
       <concept_significance>100</concept_significance>
       </concept>
 </ccs2012>
\end{CCSXML}

\ccsdesc[500]{Human-centered computing~Ubiquitous and mobile devices}
\ccsdesc[300]{Computer systems organization~Embedded systems}
\ccsdesc[100]{Computing methodologies~Neural networks}

\keywords{Time series, computing methodologies, human activity recognition, edge AI}


% Not for conference but Jounal to print: Received ...; revised ...; accepted ...

\maketitle

%%%%%%%%%%%%%%%%%%%%%%%%%%%%%%%%%%%%%%%%%%%%%%%%%%%%%%%%%%%%%%%%
\section{Introduction}

Wearable sensor-based human activity recognition has attracted significant research interest due to applications in healthcare \cite{zheng2017design, xu2018geometrical}, sports analytics \cite{zhou2022quali, singh2024novel}, smart homes \cite{bianchi2019iot}, and industrial safety monitoring \cite{tao2018worker, suh2023worker, bello2024tsak}. These systems use inertial sensors such as accelerometers and gyroscopes to recognize human activities from multivariate time-series signals, enabling applications in personalized health monitoring and assisted living.

Deploying HAR models on wearable devices remains challenging due to limited memory, computational capability, and battery capacity \cite{muhoza2023power}. Many high-performing HAR models require substantial computational resources, limiting their practicality for real-time on-device inference \cite{ronald2021isplinception}. As a result, improving hardware efficiency while maintaining strong recognition performance has become increasingly important.

Recent lightweight HAR models such as TinierHAR \cite{bian2025tinierhar}, TinyHAR \cite{zhou2022tinyhar}, and MLP-HAR \cite{zhou2024mlp} have reduced model complexity while maintaining competitive accuracy. However, many approaches still rely on recurrent temporal modeling (RNN) or computationally intensive feature extraction. Although recurrent networks such as GRU and LSTM model temporal dependencies, their sequential computation limits parallelization and can introduce latency and energy overhead on resource-constrained hardware \cite{ordonez2016deep,lin2020mcunet}.

To address these limitations, we propose LITEWAY, a fully convolutional HAR framework for efficient on-device inference. LITEWAY uses only convolutional layers and a single linear layer, achieving low computational cost and strong hardware efficiency while maintaining competitive recognition performance.

This paper makes the following contributions.
\begin{enumerate*}[label=(\roman*)]

\item We propose LITEWAY, a fully convolutional HAR framework that replaces RNN with structured convolutional decomposition, enabling efficient temporal modeling and being resource-aware.

\item We introduce a lightweight architecture optimized for wearable HAR, balancing memory, compute, and representation capacity via modular convolutional blocks.

\item Evaluated on 16 datasets, LITEWAY achieves competitive macro F1, with the Light and Full variants reducing model size by $4.06\times$--$9.52\times$ and $3.87\times$--$9.07\times$, respectively, compared to TinierHAR and TinyHAR.

\item Ablation and deployment show LITEWAY Light maximizes efficiency with $2.29\times$--$3.14\times$ lower energy, while LITEWAY Full improves macro F1 with $1.46\times$--$2.01\times$ lower energy, both versus TinierHAR and MLP-HAR, highlighting the accuracy–efficiency trade-off.

\end{enumerate*}

%%%%%%%%%%%%%%%%%%%%%%%%%%%%%%%%%%%%%%%%%%%%%%%%%%%%%%%%%%%%%%%%
\section{Related Work}
\label{sec:relatedwork}

\paragraph{Deep HAR for Wearable Sensing}

Deep learning has become the dominant approach for wearable HAR due to its ability to learn discriminative representations from multimodal sensor streams. Early architectures such as DeepConvLSTM (DCL)~\cite{ordonez2016deep} combined convolutional layers for local feature extraction with recurrent layers for sequence modeling, establishing a widely adopted \textit{CNN-RNN} paradigm for inertial sensing applications.

%Despite their effectiveness, recurrent architectures require sequential computation, limiting parallelism and increasing inference latency on wearable and embedded devices with constrained memory, compute, and power budgets.
Despite their effectiveness, recurrent architectures require sequential computation, limiting parallelism and increasing inference latency on resource-constrained wearable devices. To address these limitations, several works explored convolution-based temporal modeling approaches for sequence processing~\cite{bai2018empirical,van2016wavenet}, demonstrating strong performance while enabling low latency.

However, capturing long-range temporal dependencies through deeper or dilated convolutions may still increase computational cost and memory usage, particularly on resource-constrained wearable platforms~\cite{zhou2022tinyhar}. Consequently, efficient long-range temporal modeling remains a key challenge for real-time HAR.

\paragraph{Lightweight HAR Architectures}

Several studies have explored efficient HAR models for on-device inference. TinyHAR~\cite{zhou2022tinyhar} proposed a lightweight architecture optimized for edge deployment by reducing computational complexity through efficient convolutional operations and compact feature extraction modules. 

Beyond reducing parameter count, lightweight HAR research has focused on balancing temporal modeling capability with deployment efficiency. For example, TinierHAR~\cite{bian2025tinierhar}, SPECTRA~\cite{gurung2026spectra} and MLPHAR~\cite{zhou2024mlp} further reduced model complexity while preserving competitive performance. However, existing approaches often rely on CNN--RNN architectures. Although MLPHAR does not use recurrent modules, it is not fully end-to-end learnable.

Although prior lightweight architectures improve efficiency, existing methods still face challenges in jointly optimizing temporal receptive field, inference latency, and model compactness for streaming wearable HAR. These limitations motivate the design of convolution-only architectures that provide efficient temporal modeling while remaining suitable for low-power HAR deployment.

%%%%%%%%%%%%%%%%%%%%%%%%%%%%%%%%%%%%%%%%%%%%%%%%%%%%%%%%%%%%%%%%%%%
\section{Methodology}
\label{sec:method}

\begin{figure*}
    \centering
    \includegraphics[width=\linewidth]{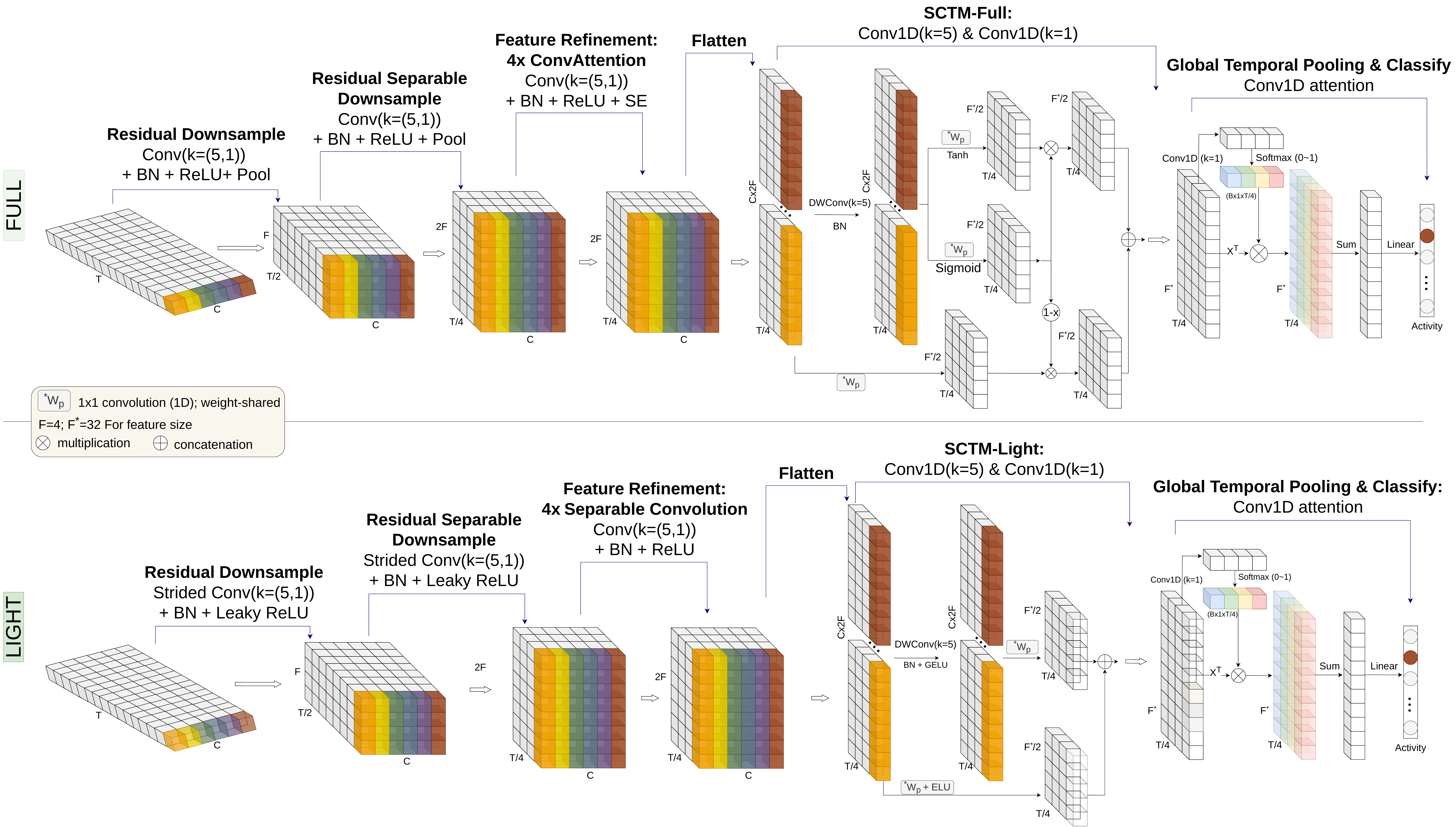}
    %\vspace{-15pt}
    \caption{LITEWAY architectures (Full and Light), based on a fully convolutional design with a single linear classification layer}
    % \caption{LITEWAY architectures (Full and Light). Both employ a fully convolutional backbone followed by a single linear classification layer.}
    \Description{Proposed Architectures for HAR. Full and its lightweight variant for edge devices, showing reduced layers and computational components compared to a standard model.
    Full model is composed by Res-Down $\rightarrow$ LightSE $\rightarrow$ SCTM-Full $\rightarrow$ ConvAtt.
    The LITEWAY Light blocks integrate StrideConv $\rightarrow$ NoSE $\rightarrow$ SCTM-Light $\rightarrow$ ConvAtt.}
    \label{fig:architecture}
\end{figure*}

We propose LITEWAY (Figure \ref{fig:architecture}), a fully convolutional architecture for time-series classification on resource-limited devices. The model has three components: 
\begin{enumerate*}[label=(\roman*)]
\item a feature extraction backbone that downsamples and refines features using residual and depthwise-separable convolutions with attention-based channel recalibration;
\item a Structured Convolutional Temporal Modeling (SCTM) module capturing long-range dependencies via depthwise convolutions, shared projections, and gated pathways without recurrence;
\item a lightweight classification head that aggregates features via attention-based pooling followed by a linear layer.
\end{enumerate*}
% Each component minimizes redundant parameters while preserving representational capacity.
Each component minimizes redundant parameters while preserving capacity.

\subsection{Convolutional Feature Extraction Backbone}

The goal of the backbone is to reduce temporal resolution early while avoiding expensive feature transformations and refinement. The backbone consists of six convolutional blocks in two stages. 

\textit{Step 1: Temporal Downsampling.} The first stage uses two residual blocks with batch norm and leaky-relu for temporal downsampling. The first block applies residual depthwise convolution, while the second block performs depthwise separable convolution. Each block is followed by pointwise mixing. This design enables efficient temporal downsampling.

\textit{Step 2: Feature refinement.}
The subsequent four layers operate on reduced temporal resolution. These blocks use depthwise separable convolutions to decouple temporal filtering from channel mixing while preserving representational capacity. Each block also includes a squeeze-and-excitation (SE) submodule \cite{hu2018squeeze,roy2018concurrent}. We implement SE using $1 \times 1$ convolutions for channel attention to maintain a fully convolutional design.

For the efficient variant, we replace conventional convolution and pooling with strided depthwise convolutions (StrideConv), which combine feature extraction and downsampling to reduce MACs. Consequently, we remove the SE modules at the cost of a tolerable performance loss.

Overall, this backbone establishes an early and sustained reduction in feature dimensionality, forming the foundation for a compact model design.

\subsection{SCTM}     % {Structured Convolutional Temporal Modeling}
SCTM is the core component of LITEWAY and integrates efficient feature transformation principles from gated architectures and multi-branch convolutional designs. Rather than introducing a new gating operation, SCTM combines these concepts into a wearable-oriented temporal modeling module optimized for HAR efficiency.% accuracy and resource efficiency -> HAR efficiency.

\subsubsection{SCTM-Full}

% --larger--
This block draws on and unifies several established design principles into a single parameter-efficient module. Given an input $X \in \mathbb{R}^{B \times C \times T}$, the block first applies a depthwise temporal convolution followed by a pointwise activation, $H = \phi(\mathrm{DWConv}(X))$, decoupling temporal filtering from channel mixing in the spirit of depthwise separable convolutions~\cite{howard2017mobilenets}, which have been shown to approximate full convolutions at a fraction of the parameter cost. A shared pointwise projection $Z = W_p(H)$ then produces a single latent representation reused across both pathways, avoiding the parameter duplication inherent in standard two-branch designs such as the Gated Linear Unit~\cite{dauphin2017language}, where two independent projections $W_1, W_2$ are learned. The block then constructs two complementary signals. The first, $Y_f = \sigma(Z) \odot \tanh(Z)$, is a collapsed Gated Tanh Unit~\cite{van2016wavenet,oord2016pixel} in which the filter and gate weights are tied to the same projection; the sigmoid acts as a soft content gate, selecting which features to pass, while the tanh provides a bounded, zero-centered nonlinear transformation, a combination empirically shown to outperform rectified activations for sequential and audio modeling~\cite{van2016wavenet,dauphin2017language}. The second signal, $Y_b = (1 - \sigma(Z)) \odot W_p(X)$, applies the complement of the same gate to a projection of the raw input, directly instantiating the carry gate of Highway networks~\cite{srivastava2015training}, where $C = 1 - T$ was introduced to allow unimpeded information flow through deep networks. Critically, the gate is derived from $Z$, which encodes temporal structure via the preceding depthwise convolution, rather than from the raw input as in the original highway formulation; the gating decision is therefore informed by processed temporal features rather than channel statistics alone. Similarly, the carry stream applies $W_p$ to $X$ rather than bypassing it as an identity, ensuring that even the preserved pathway undergoes channel mixing, preserving the complementary relationship between the two streams while maintaining a shared projection. The two streams are fused by channel-wise concatenation, $Y = \mathrm{Concat}(Y_f, Y_b)$, rather than by the addition used in highway networks~\cite{srivastava2015training} and residual connections~\cite{he2016deep}. Additive fusion combines transformed and carried information into a single representation, whereas concatenation preserves both streams separately and allows subsequent layers to learn their interaction. This design is motivated by the split-transform-merge strategy employed in Inception architectures~\cite{szegedy2015going,ronald2021isplinception}, where projected representations are processed independently before concatenation. Similarly, SCTM applies complementary transformations within parallel pathways and merges them through concatenation, increasing local representational capacity without requiring multiple full-dimensional transformations. Together, these decisions yield a block that performs temporal modeling through a single depthwise convolution and a single shared projection, without recurrent state, without separate branch weights, and without additive fusion losses, offering a principled reduction in parameter count relative to both recurrent models~\cite{chung2014empirical} and standard gated convolutional baselines~\cite{dauphin2017language}.

\subsubsection{SCTM-Light}  % {Lightweight Temporal Variant (SCTM-Light)}

The lightweight variant of SCTM that reduces computational cost. Like the full block, it applies a depthwise temporal convolution followed by a pointwise projection with GELU activation. To further reduce MACs, a residual pathway is projected using ELU and concatenated with the main stream, yielding a compact yet expressive output. This design retains a compressed residual shortcut for the input while omitting separate gate multiplication, simplifying computation while preserving complementary feature flow. Formally, 
$Y = \mathrm{Concat}\big(Z, \mathrm{ELU}(W_{\mathrm{p}}(X))\big)$, 
where $Z$ is the GELU-activated projection of the depthwise convolution.

\subsection{Global Temporal Pooling and Classification}

We aim to aggregate temporal features without introducing additional heavy sequence modeling.

\begin{enumerate*}[label=Step \arabic*:]
\item We use \textit{attention-based temporal pooling} with a single learnable projection that computes importance weights over time steps i.e. $\alpha = \mathrm{softmax}(W_a X)$.

\item \textit{Aggregation.}
The final representation is a weighted sum of temporal features, producing a compact global embedding.

\item \textit{Classification.}
A single linear layer maps this embedding to output classes, ensuring minimal parameter overhead in the decision stage.
\end{enumerate*}

\subsection{Efficiency design choices}   % {Designing an Efficient Network}

Beyond architectural design, efficiency is enforced through systematic reduction of redundant computation, especially when parameter optimization is sensitive.
\begin{enumerate*}[label=(\roman*)]
\item \textbf{Strided Convolutions:} Replace pooling operations to eliminate redundant layers.
\item \textbf{1D Convolution over Recurrent Models:} Avoid sequential hidden-state computations.
%\item \textbf{Lightweight Activations:} Use LeakyReLU instead of expensive nonlinearities.
\item \textbf{Lightweight Activations:} Prefer lightweight activation where appropriate.
\item \textbf{Selective Residual Connections:} Apply only where optimization stability requires it.
\end{enumerate*}
% \textbf{As an outcome} these choices collectively ensure that parameter and computation efficiency is maintained throughout the entire architecture.
\textbf{Collectively}, these design choices ensure parameter and computational efficiency throughout the entire architecture.

%%%%%%%%%%%%%%%%%%%%%%%%%%%%%%%%%%%%%%%%%%%%%%%%%%%%%%%%%%%%%%%%%%
\section{Experimental Results}
\label{sec:results}

\subsection{Experiment Setup}

\begin{table}[t]
\centering
% \caption{Summary of evaluated HAR datasets. In the table, $\#Subj$ denotes the number of subjects; $\#Cls$, the number of activity classes; $Ch$, the number of sensor channels; $F$ (Hz), the sampling frequency; and $SW$, the sliding window in seconds.}
\caption{Summary of the evaluated HAR datasets. $\#Subj$ denotes the number of subjects, $\#Cls$ the number of activity classes, $Ch$ the number of sensor channels, $F$ (Hz) the sampling frequency, and $SW$ the sliding-window in seconds.}
\footnotesize
\setlength{\tabcolsep}{3pt}
\begin{tabular}{llccccc}
\toprule
\textbf{Dataset} & \textbf{Sensor} & \textbf{\#Subj} & \textbf{\#Cls} & \textbf{Ch} & \textbf{$F$(Hz)} & \textbf{SW} \\
\midrule
Dg~\cite{bachlin2009potentials}        & Acc                     & 10 &  9 &  9  &  64 & 1 \\
Uschad~\cite{zhang2012usc}             & Acc/Gyro                &  7 & 12 &  6  & 100 & 1 \\
Skodar~\cite{zappi2008activity}        & Acc                     &  1 & 10 & 30  &  33 & 4 \\
Pamap2~\cite{reiss2012introducing}     & Acc/Gyro/Mag            &  9 & 12 & 18  &  33 & 4 \\
Dsads~\cite{altun2010comparative}      & Acc/Gyro/Mag            &  8 & 19 & 45  &  25 & 4 \\
Hapt~\cite{reyes2016transition}        & Acc/Gyro                & 10 & 12 &  6  &  50 & 2.56 \\
Rw~\cite{sztyler2017position}          & Acc                     & 15 &  8 & 21  &  50 & 4 \\
%WISDM~\cite{kwapisz2011activity}       & 1$\times$Acc            & 36 &  6 &  3  &  20 & 4 \\
Oppo~\cite{roggen2010collecting}       & IMU/Mag/Quat            &  4 & 18 & 77  &  30 & 4 \\
Oppoloc~\cite{roggen2010collecting}    & IMU/Mag/Quat            &  4 & 6  & 77  &  30 & 4 \\
Recgym~\cite{bian2022contribution}     & Acc/Gyro/Cap            & 10 &  7 & 12  &  20 & 4 \\
MotionSense~\cite{malekzadeh2019mobile}& Acc/Gyro                & 24 & 12 &  6  &  50 & 4 \\
Mhealth~\cite{banos2014mhealthdroid}   & IMU/ECG                 & 10 & 12 & 23  &  50 & 4 \\
Sho~\cite{shoaib2014fusion}            & IMU/LAcc                & 10 &  7 & 60  &  50 & 4 \\
Uci~\cite{Anguita2013APD}              & Acc/Gyro/LAcc           & 30 &  6 &  9  &  50 & 2.56 \\
Realdisp~\cite{roggen2010collecting}   & Acc/Gyro/Mag/Quat       & 17 & 33 & 81  &  50 & 4 \\
Wear~\cite{bock2023wear}               & Acc                     & 22 & 19 & 12  &  50 & 4 \\
\bottomrule
\end{tabular}
\label{tab:har_datasets}
% \vspace{-10pt}
\end{table}

\paragraph{Datasets and Preprocessing}
We evaluate the proposed method on 16 widely used HAR datasets covering diverse sensing modalities, sampling frequencies, and activity types (Table~\ref{tab:har_datasets}). All sensor signals are segmented using dataset-specific sliding windows with 50\% overlap. Each sensor channel is independently standardized using the mean and standard deviation computed from the training set only. Oppo and oppoloc share data but use different labels.

\paragraph{Evaluation Protocol}
We follow a subject-independent evaluation protocol for robust results. For most datasets, Leave-One-Subject-Out (LOSO) cross-validation is used to assess generalization to unseen users. For large-scale datasets (MotionSense and uci), group-based subject hold-out is adopted to reduce training time. An exception is skodar, which contains a single subject; thus, Leave-One-Session-Out is used. % to evaluate temporal generalization across sessions.

\paragraph{Training and Metrics}
All experiments are conducted on NVIDIA RTX 3090 GPU. To ensure reproducibility and reduce variance due to random initialization, each experiment is repeated with five random seeds (1--5), and the average performance is reported.
Models are trained for up to 150 epochs using the AdamW optimizer with cross-entropy loss. The initial learning rate of $1 \times 10^{-3}$ is reduced by a factor of 0.1 if no improvement is observed for 7 epochs. Early stopping is applied with a patience of 15 epochs.
We report Macro-F1 ($F1_M$) as the primary performance metric due to class imbalance across datasets. In addition, model efficiency is evaluated using the number of parameters ($nP$) and MACs (multiply--accumulate operations), enabling a direct accuracy--efficiency trade-off assessment.

\paragraph{Baselines}
We compare the proposed method against representative HAR models, including TinierHAR, TinyHAR, and MLP-HAR as state-of-the-art efficient models, as well as DeepConvLSTM, the most commonly reported efficient baseline. All baselines are evaluated under the same training and evaluation protocol to ensure a fair comparison.
% We compare the proposed method against representative HAR models, including TinierHAR, TinyHAR, and MLP-HAR as state-of-the-art efficient models, and DeepConvLSTM, the most commonly reported baseline. All baselines are evaluated under the same training and evaluation protocol to ensure fair comparison.

%%%%%%%%%%%%%%%%%%%%%%%%%%%%%%%%%%%%%%%%%%%%%%%%%%%%%%%%%%%%%%%%%%
% --------------------------------------------------
\subsection{Experimental Results}

\subsubsection{Per-Dataset Performance}

\begin{figure*}[ht]
    \centering
    \includegraphics[width=\linewidth]{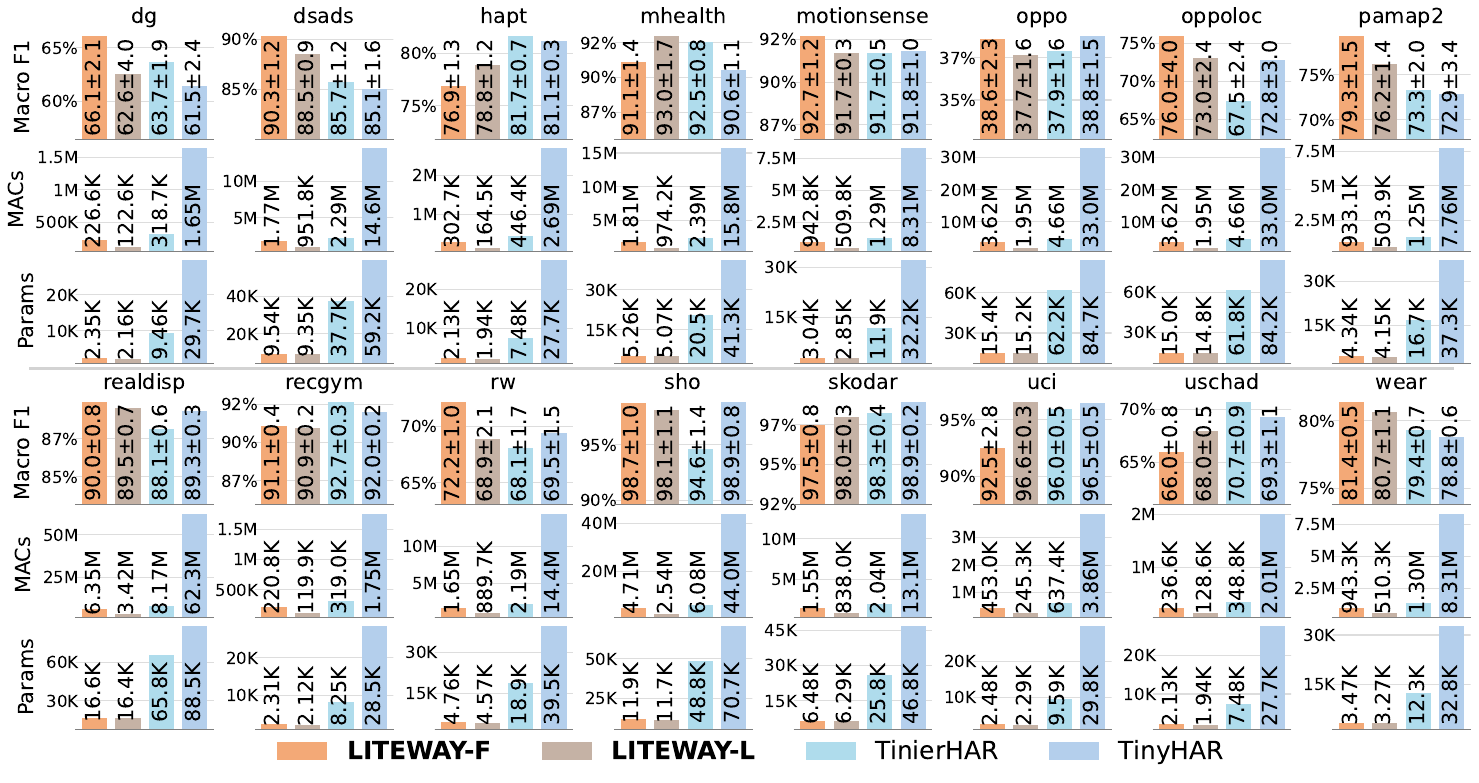}
    % \vspace{-20pt}
    \caption{Performance of LITEWAYs in terms of (1) macro F1 score, (2) MACs, and (3) number of parameters, compared with TinierHAR, and TinyHAR. The average across 16 datasets is shown in Figure~\ref{fig:overall}.}
    \Description{Experimental results comparing LITEWAYs with other state-of-the-art methods across 16 datasets. Bars show macro F1 scores, MACs, and parameter counts. Values are annotated inside the bars.}
    \label{fig:results}
\end{figure*}

We first evaluate all methods across 16 datasets to assess generalization ability. Figure~\ref{fig:results} reports detailed results for each dataset in terms of (1) macro F1 score, (2) model complexity (MACs), and (3) number of parameters.

The LITEWAY Light variant achieves top-2 macro F1 scores on 9 out of 16 datasets. It ranks first on mhealth, motionsense, and recgym, and achieves second place on dsads, hapt, pamap2, realdisp, sho, and skodar. Similarly, the LITEWAY Full variant reaches top-2 performance on 10 datasets, securing first place on dg, motionsense, and pamap2, and second place on dsads, hapt, mhealth, realdisp, rw, sho, and skodar. These results demonstrate that both LITEWAY variants consistently rank near the top despite using substantially fewer parameters and MACs than larger models such as TinyHAR.

Even on more challenging datasets such as oppo and oppoloc, where macro F1 scores are lower across all methods, LITEWAY Light and Full maintain competitive rankings, typically second or third. This highlights their robustness and strong generalization ability under diverse and difficult conditions. Overall, these findings indicate that the proposed methods provide a favorable balance of high accuracy and efficiency across all evaluated datasets.

\textbf{Finding:} Both LITEWAYs achieve comparable performance across divers datasets, demonstrating that our proposed methods deliver strong generalization while remaining lightweight and efficient.

% --------------------------------------------------
\subsubsection{Overall Performance}

\begin{figure}[t]
    \centering
    \includegraphics[width=\columnwidth]{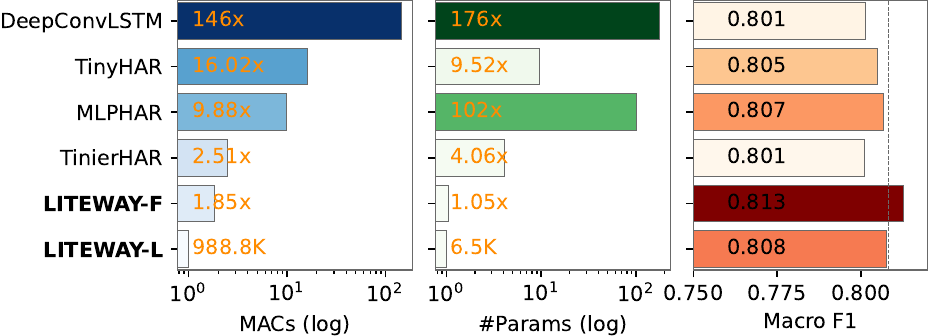}
    % \vspace{-10pt}
    \caption{Comparison of macro F1, MACs, and parameters.}
    % \caption{Comparison of macro F1, MACs, and parameters averaged over 16 datasets.}
    % \caption{Comparison of macro F1-score, MACs, and number of parameters across models.}
    \Description{Results summary}
    \label{fig:overall}
    % \vspace{-10pt}
\end{figure}

Figure~\ref{fig:overall} summarizes the average performance across all 16 datasets, reporting macro F1-score alongside relative computational cost (MACs) and model size (parameters).

Among all methods, \textit{LITEWAY Full} achieves the highest macro F1-score (0.813), followed by \textit{LITEWAY Light} (0.808). Both models outperform existing baselines, including TinierHAR (0.801), TinyHAR (0.805), MLPHAR (0.807) and DeepConvLSTM (0.801).

In terms of efficiency, \textit{LITEWAY Light} requires the lowest cost (988.8K MACs) and model size (6.5K parameters), while \textit{LITEWAY Full} maintains a similarly compact footprint with only a small increase in complexity. Compared to existing baselines, this corresponds to reductions of approximately $2.51\times$--$146\times$ in MACs and $4.06\times$--$176\times$ in parameters, highlighting the substantially lower resource requirements of the proposed designs.

\textbf{Finding:} LITEWAY variants achieve competitive macro F1, while \textit{LITEWAY Light} sets the lowest model size and MAC.
% \textbf{Finding:} The proposed models achieve the best overall macro F1 scores, with \textit{LITEWAY Light} establishing a new lower bound in model size and MAC.

% --------------------------------------------------
\subsubsection{Trade-off between Efficiency and Performance}

We analyze the relationship between efficiency and recognition performance, as shown in Figure~\ref{fig:overall}.

Two key trends emerge: (i) increasing model complexity does not consistently improve accuracy, with gains often saturating or remaining limited despite higher MACs and parameter counts, and (ii) parameter count is a poor proxy for computational cost, as models with similar sizes can exhibit substantially different MACs.

In contrast, the proposed models operate in a more efficient regime. \textit{LITEWAY Light} represents the extreme low-cost setting, while \textit{LITEWAY Full} achieves higher accuracy with only marginal additional cost, indicating better utilization of model capacity.

% \textbf{Finding:} The proposed architectures lie on the Pareto frontier, achieving comparable accuracy while remaining hardware efficient.
\textbf{Finding:} The proposed networks achieve substantially lower energy consumption without compromising recognition performance.

% --------------------------------------------------
\subsection{Ablation Study}

%---
\subsubsection{Ablation setup}

The impact of key architectural components was evaluated using the following ablations.
Three residual configurations: \textbf{Res-Down} (default, residuals in early layers), \textbf{Res-All} (residuals in all layers), and \textbf{NoRes} (no residuals). 

Channel recalibration was studied via \textbf{LightSE}, an efficient variant of squeeze-and-excitation that reduces computation while providing a middle ground between SE and no-SE (\textbf{NoSE}).  

Aggregation strategies were compared, including \textbf{ConvAtt} (proposed convolutional attention), \textbf{LinAtt} (linear attention, similar to TinyHAR and TinierHAR), and \textbf{MMX} (max-mean pooling), to assess their influence on accuracy and efficiency.

% Finally, activation strategies were compared, including homogeneous GELU, homogeneous Leaky ReLU, GELU$\rightarrow$Leaky (uniform replacement of GELU by Leaky ReLU), and a heterogeneous design in LITEWAY Light, where different blocks use ReLU, Leaky ReLU, or GELU.
Finally, activation strategies were compared, including homogeneous GELU, homogeneous Leaky ReLU, GELU$\rightarrow$Leaky (uniform replacement of GELU by Leaky ReLU), and a heterogeneous design in LITEWAY Light, where blocks use ReLU, Leaky ReLU, or GELU.

This setup allows us to quantify the contributions of residual connections, attention, and activation functions to both accuracy and efficiency.

%---
\subsubsection{Design Ablation}

\begin{table}[t]
\scriptsize
\centering
\caption{Ablation of LITEWAY variants; each row shows a single architectural modification. Empty cells indicate the default modules used in LITEWAY Full.}
% \caption{Ablation of LITEWAY variants; each row shows a single architectural modification, with empty cells indicating default modules of LITEWAY Full.}
\resizebox{\columnwidth}{!}{
% \begin{tabular}{l@{\hspace{0.8em}}cccc@{\hspace{1.2em}}rrr}
\begin{tabular}{l|llll|ccc}
\toprule
              & Downsample  & Refine & Temporal & Aggregate   & $F1_M$   & nP     & MAC  \\
\midrule
\textbf{LITEWAY-F}     & Res-Down & SE     & SCTM-F  & ConvAtt     & 81.3 & 6.7K   & 1.8M \\
\midrule
NoRes         & NoRes       &        &          &             & 80.6 & 6.6K   & 1.6M \\
StrideConv    & StrideConv  &        &          &             & 80.7 & 6.7K   & 1.3M \\
LightSE       &             & LightSE   &          &             & 80.6 & 6.6K   & 1.8M \\
NoSE          &             & NoSE   &          &             & 80.9 & 6.5K   & 1.8M \\
SCTM-L        &             &        & SCTM-L   &             & 80.6 & 6.7K   & 1.6M \\
LinAtt        &             &        &          & LinAtt      & 80.8 & 6.7K   & 1.8M \\
MMX           &             &        &          & MMX         & 79.7 & 6.7K   & 1.8M \\
\midrule
\textbf{LITEWAY-L}     & StrideConv  & NoSE   & SCTM-L   & ConvAtt     & 80.8 & 6.5K   & 989K \\
\bottomrule
\end{tabular}
}
\label{tab:ablation_layers}
% \vspace{-10pt}
\end{table}

Table~\ref{tab:ablation_layers} evaluates the impact of LITEWAY architectural blocks on performance and efficiency. Parameter counts are similar across variants (6.7–6.9K), but MACs vary significantly. LITEWAY Full achieves the highest $F1_M$ (81.3), serving as the performance reference. Removing or simplifying components generally reduces $F1_M$: NoRes and SCTM-L decrease $F1_M$ by 0.7 points, while MMX achieves the lowest $F1_M$ (79.7), highlighting the importance of ConvAtt aggregation. LITEWAY Light achieves a favorable trade-off, reducing MACs by over 45\% relative to the full model, with only a minor $F1_M$ drop (from 81.3 to 80.8).

\textbf{Finding:} Residual connection, SCTM, and ConvAtt are critical for high $F1_M$, while efficient blocks allow LITEWAY Light to maintain strong performance with minimal resource demand.

\begin{table}[t]
\centering
\scriptsize

\begin{minipage}[t]{0.48\linewidth}
    \centering
    \begin{tabular}{l|ccc}
    \toprule
    Model     & $F1_M$ & nP   & MACs \\
    \midrule
    NoRes       & 80.1 & 6.4K & 904K \\
    Res-All     & 80.0 & 6.8K & 1.3M \\
    LITEWAY-L   & 80.8 & 6.5K & 989K \\
    \bottomrule
    \end{tabular}
    
    \captionof{table}{Effect of residual connections (16 datasets).}
    \label{tab:residual}
\end{minipage}
\hfill
\begin{minipage}[t]{0.48\linewidth}
    \centering
    \begin{tabular}{lccc}
    \toprule
    Model                    & MAC  & nP   & F1 \\
    \midrule
    GELU                     & 1.4M & 6.5K & 80.6 \\
    Leaky ReLU               & 1.3M & 6.5K & 80.5 \\
    GELU $\rightarrow$ Leaky & 978K & 6.5K & 80.4 \\
    LITEWAY-L                & 989K & 6.5K & 80.8 \\
    \bottomrule
    \end{tabular}
    
    \captionof{table}{Effect of activation functions (16 datasets).}
    \label{tab:activateions}
\end{minipage}
% \vspace{-10pt}

\end{table}

%---
\subsubsection{Activation function}

We evaluate activation strategies (Table~\ref{tab:activateions}) considering homogeneous GELU, homogeneous Leaky ReLU, GELU$\rightarrow$Leaky replacement, and the heterogeneous design in LITEWAY. Homogeneous ReLU was excluded as it is less accurate.

Activation functions have limited impact on accuracy ($\leq 0.2$ F1 variation) but affect computation. Leaky ReLU lowers MACs compared to GELU, and GELU$\rightarrow$Leaky reduces computation from 1.4M to 978K MACs without changing parameter count. The proposed LITEWAY Light achieves the best trade-off, obtaining the highest macro F1 score (80.8) with low cost (989K MACs). This indicates that block-wise assignment of activation functions is more effective than using a single activation throughout the network.

\textbf{Finding:} Mixed activations yield better accuracy–efficiency balance than uniform designs. 

%---
\subsubsection{Effect of Residual Connection}

We evaluate residual connections using three configurations: NoRes, Res-All, and LITEWAY (Res-Down). Table~\ref{tab:residual} summarizes results across 16 datasets.

NoRes achieves the lowest execution cost (904K MACs) with slightly fewer parameters while maintaining competitive performance (80.1 F1). Res-All increases computation (1.3M MACs, 6.8K params) without improving accuracy (80.0 F1), indicating limited benefit from applying residuals uniformly across all layers in lightweight HAR models.

In contrast, LITEWAY achieves the best trade-off (80.8 F1) with near-NoRes cost (989K MACs, 6.5K params), showing that residual connections are most effective when applied only in early layers.

\textbf{Finding:} Selective residual connections in early layers are sufficient for stable optimization in lightweight HAR models.

%---
\subsubsection{Deployment on Hardware}
\begin{table}[t]
  \centering
  \scriptsize
  \caption{Comparison of SOTA and LITEWAY on STM32L4S5}
  \def\al{90}
  \begin{tabular}{lrrrrrr}
    \toprule
    Model &
    \rotatebox[origin=c]{\al}{\shortstack{Inf. Time\\(ms)}} &
    \rotatebox[origin=c]{\al}{\shortstack{Weight\\(KiB)}} &
    \rotatebox[origin=c]{\al}{\shortstack{Activation\\(KiB)}} &
    \rotatebox[origin=c]{\al}{\shortstack{Cycles\\/MAC}} &
    \rotatebox[origin=c]{\al}{\shortstack{CPU\\(\% load)}} &
    \rotatebox[origin=c]{\al}{\shortstack{Energy\\(mJ/Inf)}} \\
    \midrule
    TinyHAR & $249.01 \pm 0.07$ & 107.48 & 62.70 & 13.51 & 24 & $19.14 \pm 1.03$\\
    MLPHAR & $114.81 \pm 0.02$ & 342.41 & 40.54 & 11.99 & 11 & $8.73 \pm 0.39$\\
    TinierHAR & $81.42 \pm 0.01$ & 39.54 & 14.43 & 25.58 & 8 & $6.36 \pm 0.14$\\
    \midrule
    \textbf{LITEWAY-F} & $56.71 \pm 0.00$ & 10.63 & 16.35 & 33.86 & 5 & $4.35 \pm 0.21$\\
    \textbf{LITEWAY-L} & $37.44 \pm 0.00$ & 10.07 & 16.03 & 30.54 & 3 & $2.90 \pm 0.16$\\
    \bottomrule
  \end{tabular}
  \label{tab:model-comparison}
\end{table}

To evaluate our models on edge devices, we deploy them on the low-power STM32L4S5 microcontroller running at 120\,MHz). Power is measured with the ST X-NUCLEO-LPM01A shield. Average inference time is recorded over 16 cycles via the system clock and UART, and energy is measured after disabling all GPIOs.
Table~\ref{tab:model-comparison} summarizes the hardware efficiency metrics, including inference latency, memory footprint, CPU utilization, and energy consumption. 

Among all evaluated methods, LITEWAY Light achieves the best overall efficiency on the STM32L4S5, requiring only 37.44\,ms per inference, 3\% CPU load, and 2.90\,mJ per inference. Compared with TinierHAR, it reduces both latency and energy by more than 2$\times$, while using only 10.07\,KiB of weights, corresponding to an approximately 4$\times$ smaller parameter footprint.
Although LITEWAY has a higher cycles/MACC than some baselines, its substantially lower MAC count (Figure \ref{fig:overall}) leads to lower total execution cost.

\textbf{Finding:} LITEWAY achieves superior embedded deployment efficiency across latency, memory, CPU, and energy. %, offering better hardware-performance trade-offs.

% --------------------------------------------------
\subsection{Discussion Summary and Future Work}
\label{sec:discussion}

\paragraph{Generalization}
LITEWAY models show consistent performance across diverse HAR datasets with varying modalities and complexities. Even on more challenging datasets (e.g., dg, oppo), they remain competitive, demonstrating robust feature learning despite their lightweight design. Following the Bayesian analysis of classifier comparisons advocated by
Benavoli et al.~\cite{benavoli2017time}, we ran the Bayesian signed-rank test~\cite{benavoli2014bayesian} on the per-dataset macro-F1 scores across the 16 datasets, using a region of practical equivalence (ROPE) of one F1 point.
For LITEWAY-F, the posterior probability of outperforming each
baseline ranges from $0.59$ to $0.86$ and never favours a baseline (all $P(\text{baseline better}) \le 0.13$); since no comparison crosses the conventional $0.95$ decision threshold, we conclude that our architecture is at least on par with, and most likely superior to, the SOTA baselines while remaining substantially smaller.

\paragraph{Efficiency–Accuracy Trade-off}
Increasing model size does not reliably improve performance. Larger models raise cost with limited gains, while the proposed models maintain strong accuracy with far lower MACs and parameters. The low energy demand and latency of LITEWAY further confirm its superiority, placing the models on the Pareto frontier and emphasizing efficient design over scale.

\paragraph{Design Insights}
Ablation results show that (i) residual connections are most effective in early layers, (ii) attention-based aggregation improves performance over simpler methods, (iii) temporal modeling remains essential, and (iv) heterogeneous activations provide better efficiency–accuracy balance than uniform choices.

\paragraph{Limitations and Future Work}
LITEWAY was evaluated on a single microcontroller, which may limit generalization across hardware. Real-world deployment could require hardware- and data-aware adaptations. We did not explore further optimizations such as quantization, pruning, hardware-specific acceleration, or architectural choices including SCTM depth, kernel size, and multi-sensor fusion. Improving cycle/MAC efficiency is also left for future work.

%%%%%%%%%%%%%%%%%%%%%%%%%%%%%%%%%%%%%%%%%%%%%%%%%%%%%%%%%%%%%%%%%%
% Conclusion
\section{Conclusion}
\label{sec:conclusion}

% This paper presents LITEWAY, a highly efficient HAR framework that substantially reduces computation and model size. It replaces recurrent architectures with structured convolutions, enabling efficient yet expressive feature learning. Ablation shows MACs can be greatly reduced even when parameter compression is limited. Low energy use and fast inference confirm strong deployability on resource-constrained devices.
This paper presents LITEWAY, an efficient HAR framework that substantially reduces computation and model size. It replaces recurrent architectures with structured convolutions, enabling efficient, expressive feature learning. Ablation shows MACs can be reduced despite limited parameter compression. Low energy use and fast inference confirm deployability on resource-constrained devices.

\begin{acks}
This research was supported by the Carl Zeiss Stiftung, Germany, through the Sustainable Embedded AI project (P2021-02-009) and by BMFTR in the project Cross‑Act (01IW25001).
\end{acks}

\bibliographystyle{ACM-Reference-Format}
\bibliography{references}

% \appendix

\end{document}